# Deep Sensing of Urban Waterlogging

**ABSTRACT** In the monsoon season, sudden flood events occur frequently in urban areas, which hamper the social and economic activities and may threaten the infrastructure and lives. The use of an efficient large-scale waterlogging sensing and information system can provide valuable near real-time disaster information to facilitate disaster management and enhance awareness of the general public to alleviate losses during and after flood disasters. Therefore, in this study, a visual sensing approach driven by deep neural networks and information and communication technology was developed to provide an end-to-end mechanism to realize waterlogging sensing and event-location mapping. The use of a deep sensing system in the monsoon season in Taiwan was demonstrated, and waterlogging events were predicted on the island-wide scale. The system could sense approximately 2379 vision sources through an internet of video things framework and transmit the event-location information in 5 min. The proposed approach can sense waterlogging events at a national scale and provide an efficient and highly scalable alternative to conventional waterlogging sensing methods.

## I. INTRODUCTION

In recent years, increasingly severe weather phenomena have been experienced worldwide. The occurrence of urban floods has become more frequent owing to short-term locally heavy rainfall events. Heavy rainfall may lead to flooding events and incur damages in urban areas, with the damage concentrated in the highly developed plain areas [1]. Moreover, the urban areas include numerous buildings and concrete floors, which considerably increases the complexity of the terrain and reduces the drainage efficiency [2].

Even modern cities, which are protected by extensive drainage systems and large-scale pumping stations, are prone to flooding during short periods of heavy rainfall. Although the range of such flooding is usually less than a few blocks, the dense urban road network may be severely occluded, which may hamper the economic activities and residents' lives. Improving the flood control capability of the internal waters through hardware construction incurs prohibitive costs and such structures may not be adequately effective against unpredictable extreme weather. With practical experience, disaster management strategies have gradually transformed to pre-disaster prevention and disaster reduction strategies with the development of avoidance capabilities during disasters. Nevertheless, because natural disasters cannot be completely avoided, information and monitoring technology can be used to alleviate the possible impact of disasters [3, 4].

To reduce the damage caused by such flood events and provide effective disaster response and emergency plans, it is essential to quickly analyze the data collected from the disaster area [5]. There are many sources from which observational data can be collected. They include gauge data [6, 7], remote sensing data [8, 9], and field data collection [10]. Onsite field data collection methods include sending people to the disaster-stricken area to investigate and record data after the flood event.

Observational gauge data are provided by onsite gauges that apply direct physical measurements to provide real-time, accurate readings of the height and flux of water for the monitored locations. For clarity, a real-time system should return results in milliseconds to seconds, whereas a near real-time system returns results in seconds to minutes. However, such gauges are typically sparsely distributed owing to the limitations of installation location, range, maintenance, survival rate after floods, etc., resulting in extremely sparse observations. Consequently, these gauges often do not provide sufficient information to map the flooded area [11].

Remote sensing technologies such as satellite imagery and aerial imagery are also widely used for measuring the wide-area water level and scope for defining a flooded area. An advantage of this technology is that it remotely acquires data of absolute water elevations, which is helpful for the integration of flood management and for environmental science research. However, many satellites have restrictions on their orbital cycles and inter-track spacing of satellite movements, which make them less useful for immediate information and continuous monitoring [12]. Aerial imagery is another common data source, but it also depends on weather conditions and is expensive. Moreover, the relevant information of a flooded area needed from remote imagery can be easily blocked by buildings in urban areas.

The global popularity of low-cost sensors, especially vision sensors, as well as the rise of the Internet and the improvement of communication quality, has opened up the possibility of large-scale image monitoring. The utility of visual sensing has been increasingly recognized when using "cameras as sensors" to capture real-time information during and immediately after floods [5].

The visual sensing approach is widely applied in multiple domains such as surveillance, traffic, disaster management, and assisted living [13-15]. Smart cameras exhibit a broad coverage compared to that of water level sensors (i.e., ultrasonic distance sensors, pressure sensors, manhole sensors, and sewerage sensors) and can facilitate the assessment of the flood impact on the society. Traditional flood sensing models employ landmarks and rule-marks as reference objects to identify the flood extent boundaries by computing the edge and segments; however, this process is labor-intensive and time-consuming for a national-scale application [12, 16]. Although certain methods can automatically compute the intersections of the flood marks, site-specific calibration and measurements must be realized and, thus, the application of these approaches to numerous cameras is challenging [17, 18].

To achieve visual sensing on a specific flood objective from remote scene images, computer vision techniques (including machine learning methods) are traditionally used for estimating water level [19], water region [11], and the fluctuating trend [12]. However, these computer vision-based methods all need human intervention to a certain extent to define, select, extract, and analyze the proper features of objects in the sensing pipeline. In computer vision and image processing, a feature is a piece of information about the content of an image, usually about whether a certain region of the image has certain properties.

In this context, a few cameras from a small district through a small internet of video things (IoVT) or internet of cameras (IoC) framework can be operated efficiently via a video monitoring center. However, this practice requires specific arrangements of space and equipment and on-duty staff, which may not be available at the island-wide scale [11]. Consequently, it is essential to ensure rapid screening of waterlogging with high automation and flexible control. Under the technological and economic constraints, the use of an intelligent visual sensing system with a large-scale capability is necessary.

To mitigate the damage caused by flood events and for effective disaster response and emergency plans, the rapid analysis of data collected from the affected area is essential. In recent years, with the rise of deep learning technology, especially in the application of image recognition, remarkable breakthroughs have been made in the academic community [20, 21]. Currently, image recognition is one of the most successful areas of deep learning. The major difference between deep learning of convolutional neural networks (CNNs) and conventional machine learning (ML) methods is that traditional ML methods rely on hand-crafted features for image classification, so the corresponding results will be profoundly affected. In particular, the designed image features are usually ideal and simple, and are limited to human choice, making them difficult to use in complex and changeable content image applications. In contrast, the deep learning covers the image features extracted by its convolutional neural network architecture, and generates a multiplicity of specific features through numerous image samples in the training phase. These trained specific features and other characteristics summarized from images in real life have more uses than those designed by human experts. Therefore, CNNs provide state of the art performance in image recognition and are suitable for end-to-end applied recognition applications [22].

Recently, CNNs have been innovatively applied in flood sensing research. In particular, a CNN model has been used to classify input camera scenes as foggy, stained or normal, to allow the visual sensing system to select an appropriate algorithm to segment the flood region [23]. Moreover, transfer learning of CNNs has been implemented to infer the water level by using the lasso regression method, and satisfactory results have been obtained [24]. In addition, a CNN can be trained with multiple inputs and fed with a flood image and relative absolute water level of the image as a pair to output the water level-ranking [25]. The water level-ranking involves mapping to the height at the centimeter level. However, these approaches may require the retraining of the neural network and performing regression for each camera. Moreover, for separate training inputs, the ground truth of the absolute water level for every image must be provided, which is usually derived from manual labeling. Thus, the scalability of this approach is limited.

Another approach to estimate the water level is to consider the immersion of ubiquitous objects, for which the height is known, as a reference. This kind of approach involves two subcategories: object detection (OD) and semantic segmentation (SE), which are based on different CNN techniques. Using the OD method, the reference objects can be detected from the videos of the flood and non-flood periods. Subsequently, the height difference between the detected reference object and standard reference object can be used to estimate the flood level [26]. The reference objects can be pavement fences, ashbins, post boxes, traffic buckets [27], people or bicycles [25]. The main limitation of these approaches is that the above-mentioned specific reference objects must be present in the video footage. Moreover, human effort is required to calibrate the height of each reference object or create an annotation for the training data and test data in advance.

In the extension method, SE is applied to detect the reference object at the pixel level, which can provide an approximation contour rather than an outlier bounding box. Moreover, this method can be used to estimate the water depth based on the reference object with a known height (e.g., bicycles [28] and tires [29]). However, such objects must be visible in the scene to extract the information. In particular, SE can be used to obtain the number of pixels in the water body instead of the reference objects to infer the trend of the flooding level [30]. This approach does not refer to the real size of the reference, which may reduce the preparation effort to a certain extent. However, the flood pixel ratio, which is used as a signal, is susceptible to factors such as the camera displacement, changes in the scene content, and objects (such as persons and vehicles) entering and leaving the scene. Similar to the previously mentioned image classification method, the SE method can be used as a flood detector [31]. However, the OD and SE processing workload is significantly higher compared with that of the image classification methods.

Three main types of approaches are involved in CNN-based flood sensing research. These approaches have obtained promising results, and although certain limitations remain, considerable progress has been attained in intelligent visual sensing. In this context, near real-time information of the flood event and transparency of information play key roles in disaster impact management and tolerance. Thus, to realize the near real-time sensing and broadcast of urban floods, it is necessary to integrate the cross-field integration of sensor networks, flood cognition methods, and information technology.

Considering these aspects, this study was aimed at developing a large-scale cognitive sensing system driven by deep learning techniques and information and communications technologies to perceive waterlogging through an IoC framework at the island-wide scale (approximately 2.4k vision sources). To promptly determine whether a flood event is occurring in the scene, a flexible strategy of visual sensing modules was used in the system. The main sensing module applies a classification model (section III.C) to rapidly screen the island-wide video source. This model does not require retraining and regress for each camera, and no additional annotation work is required except for the class label of the image itself. Then, the on-demand module (section III.D) employs the OD and SE models to grade and estimate the water level, respectively. These models share the same training dataset to diminish the effort of data preparation and use a unique universal reference to maximize the availability of reference. The key contributions of this work include the following:

(1) We integrate the deep learning approach within an end-to-end cyber-physical system in scale.
(2) We use the Internet of Cameras to conduct large-scale island-wide visual sensing experiments in real flood events (which is the first such work to the best of our knowledge).
(3) A GIS map and an instant notification are presented to handle the dynamic flood events to provide users with

TABLE I
ANALYSIS OF FLOOD SENSING APPLICATIONS AGAINST SENSING SOURCES

| Sensing source | Ref. | Year | Objective | Flood Detection | Water Leveling | Geo. Mapping | Notifying | Scope of the Study |
|---|---|---|---|---|---|---|---|---|
| Crowdsourced/social media images | [32] | 2017 | Detecting flood extent from crowdsourced images | v | v | – | – | Real-world, tested on collected dataset |
| | [33] | 2019 | Detecting flood extent on inundated roadways | v | – | – | – | Real-world, tested on collected dataset |
| | [25] | 2020 | Water level prediction from social media images | v | v | – | – | Real-world, tested on collected dataset |
| | [34] | 2020 | Quantifying flood water levels with VGI | v | v | – | – | Real-world, tested on a section of road |
| | [35] | 2020 | Screening flooding photo from social media | v | – | v | – | Real-world, tested on collected dataset |
| | [36] | 2020 | Flood severity mapping with VGI | v | v | v | – | Real-world, tested on collected dataset |
| | [37] | 2020 | Assessing flood severity from crowdsourced photos | v | v | – | – | Real-world, tested on collected dataset |
| In situ video streaming | [11] | 2015 | Detecting flood extent from live camera | v | v | – | – | Real-world, tested on one river |
| | [12] | 2015 | Monitoring water level and water level fluctuation | v | v | – | – | Real-world, tested on two rivers |
| | [38] | 2015 | Flood detection from surveillance cameras | v | – | – | – | Real-world, tested on one footage |
| | [39] | 2017 | Real-time flood detection and alert by mobile App. | v | – | – | v | Real-world, tested on one river |
| | [16] | 2018 | Flood inundation forecasts | v | v | – | – | Real-world, tested on eight sites |
| | [30] | 2019 | Monitoring flood level trend with surveillance cameras | v | v | – | – | Real-world, six surveillance footages |
| | [40] | 2019 | Roadway flood detection with PTZ cameras | v | – | – | – | Real-world, tested on one site |
| | [41] | 2020 | River state classification for flood risk exists or not | v | – | – | – | Real-world, tested on 20 sites |
| | [42] | 2021 | Estimating water level of urban streams in real time | v | v | – | – | Real-world, tested on two rivers |
| | Pro. | 2021 | System-wise end-to-end flood sensing, geo-mapping, and awareness | v | v | v | v | Real-world, tested nationwide |

Note: Pro. = system proposed in this current study, PTZ = pan-tilt-zoom, VGI = volunteered geographic information

flexible and feasible awareness when waterlogging occurs.

The remainder of this paper is organized as follows. Section II reviews the relevant literature. Section III describes the deep sensing system, including the data I/O, flood perception modules, GIS mapping and notifications, and the corresponding results. Section IV discusses possible directions for future work in the area. Finally, Section V presents concluding remarks.

## II. RELATED WORK

The focus of this work is the achievement of system-wise near-real-time end-to-end flood-mapping services. As mentioned before, using deep learning to remotely sense flood events from real-time ground-level images is still a relatively new idea, and intelligent end-to-end flood-mapping systems are also rare. In Table I, we summarize those works that are closely related to this study. As demonstrated in Table I, all related papers have sorted based on their sensing source and publication date. It also shows whether they have implemented the flood detection, water leveling, geographic mapping, and notifying or not, and we have marked them with a "v" mark and a "–" mark respectively. The last row of the table belongs to the system proposed in this current study,

### A. Sensing from Crowdsourced and Social Media Images

The convenience and versatility of smartphones have resulted in the in situ information collected from crowdsourced or crawled from social media (secondary sources) being frequently used as input data for sensing floods [32, 33].

Witherow et al. [34] proposed a flood detection and water level determination methodology that uses image processing and photogrammetric methods to process photos from smartphone cameras with volunteered geographic information (VGI) that is then linked to a local terrain model. Ning et al. [35] implemented a prototype screening system to identify flooding-related photos from social media (e.g., Twitter, Facebook, and Instagram). Feng et al. [36] also used social media posts with VGI and interpreted water level with people contained in the images as a scale rule. These crowdsourced photos from smartphones that are associated with their geographic locations can provide free, timely, and in situ visual information about flood events to decision-makers.

Pereira et al. [37] focused on flood severity assessment from crowdsourced social media photos with deep neural networks. Chaudhary et al. [25] focused on an automated method for predicting water level from social media images with pairwise ranking. However, these works focused on flood detection rather than flood mapping and further awareness services.

However, as social media photos lack quality control, the time and location of a flood event may not be correct. Furthermore, the near-real-time aspect is dependent on the immediacy with which the images are uploaded by the public. Thus, although the people in the community may be able to go deep into certain key locations and collect images from different perspectives, there may be many uncertain quality factors that affect the correctness of crowdsourced data. Moreover, the distribution range and density of the observation points cannot be ascertained in advance, which hinders the analysis results and feasibility of the overall situation.

### B. Sensing from In Situ Video Streaming

Recently, directed real-time streaming video sources (primary sources) that integrate image processing methods and deep learning with CNN have been used to analyze the live situation of flood development.

In earlier studies, Lo et al. [11], [12] combined real-time streaming video and image processing methods to automatically monitor the water level and extent of floods. In addition, the continuous video frames can also be used for this visual sensing method to clearly determine the fluctuation of the water level. Filonenko et al. [38] used background subtraction and color probability to divide the flood region on roadway images. In a live camera approach, Bhola et al. [16] used the size of bridges and the detected water surface in the images to estimate the water level. Further, Menon and Kala [39] used a region-based image segmentation method (GrowCut) to detect the flood extent and provide warning information by mobile app.

Deep CNNs are being more frequently used because of their state-of-art performance. One such approach for flood level trend monitoring with surveillance cameras was reported by Moy et al. [30], who used U-net [43] to segment the water body and compute the proportion of water-covered area over a series of video frames. Son et al. [40] also used FNC-AlexNet to segment a flood region from the pan-tilt-zoom (PTZ) cameras. Oga et al. [41] proposed a method that combines patch processing and CNN to classify the river state from images captured by a river surveillance camera.

More recently, Jafari et al. [42] proposed a real-time water level monitoring approach using images from live cameras. They applied a deep learning-based semantic segmentation algorithm to label the water body and reference objects (staff gauge, pier) as a scale. Their proposed method was verified via laboratory experiments and field experiments on two urban rivers. They combine live cameras and deep learning to achieve a timely processing flow, with primary focus on the water level of urban rivers.

Although many studies have focused on CNN-based flood sensing with the primary source being live cameras, studies that are focused on system-wise end-to-end flood-mapping from in situ IoC during floods, with near-real-time visual sensing and announcement systems are still limited.

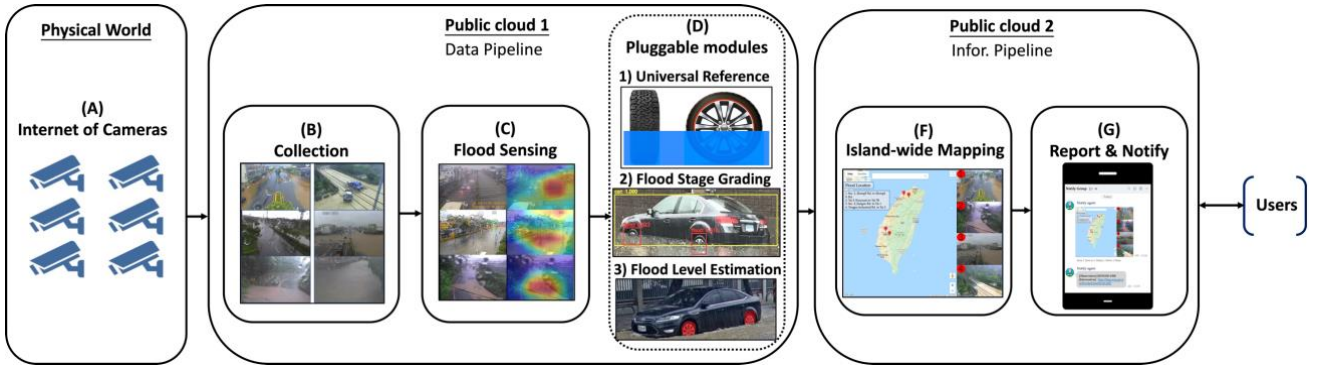

FIGURE 1. Overview of the end-to-end visual sensing system. The system acquired images from the physical world through a network of cameras and provided the situation information of the island-wide flooding to the users.

## III. DEEP SENSING SYSTEM AND CASE RESULTS

An end-to-end visual sensing system was used to intelligently and automatedly process the input from the physical world and output the information to the users. The system was expected to exhibit an enhanced data acquisition capability, access the hidden modules, and broadcast the release flooding information to users over-the-air. As shown in Fig. 1, the proposed end-to-end visual sensing system involved two main modules. The data pipeline performed the data acquisition, stream decode, and scene inference. Next, the information pipeline compiled the inference outcome and deployed the results to the information hub. Each sub-module in the pipelines could operate in a cloud resource; in this study, the two main pipeline modules were implemented in two public clouds, namely, the Taiwan computing cloud (TWCC) and Amazon Web Services to optimize the network usage and data transfer. This strategy could help retain the massive data I/O inside the independent cloud and only transfer the pruned outcomes to the subsequent cloud. The detailed function of each sub-module is described in the following subsections.

### A. Internet of Cameras

Recent developments in vision sensors have led to their widespread use in a variety of applications, such as video surveillance, traffic monitoring, crowd counting, behavior understanding, and flood detection. In Taiwan, the camera networks are distributed island-wide for different purposes. To realize traffic monitoring and ensure public safety, the cameras are distributed in proportion to the human activity density and urbanization level. Consequently, the camera density is lower in areas with a sparse population and fewer public facilities in which the waterlogging may not have a direct impact. In this study, several camera networks that covered the high human activity areas were considered, as shown in Fig. 2(a). To ensure an appropriate coverage to sense the waterlogging distribution, 2379 camera sources were used to provide the perception input of the physical world. Several

TABLE II
SPECIFICATIONS OF THE CAMERA NETWORKS.

| SOURCE* | CODEC | RESOLUTION | QUANTITY |
|---|---|---|---|
| DGH | MJPEG | 320×240<br>352×240<br>480×270<br>720×480 | 1424 |
| NTPC | MJPEG | 800×600 | 289 |
| TYC | MJPEG | 320×180<br>320×192<br>320×240<br>352×240<br>480×270<br>704×480<br>800×464 | 123 |
| TYC (SEWER) | FLV | 1280×720 | 29 |
| TNC | MJPEG | 352×240<br>704×480<br>960×480<br>1920×1080 | 148 |
| KC | JPEG | 320×240<br>352×240<br>640×480<br>704×480<br>720×480<br>1280×720<br>1280×1024 | 341 |
| NC | JPEG | 352×240<br>720×480 | 25 |
| | | | TOTAL: 2379 |

*DGH: Directorate General of Highways. NTPC: New Taipei City. TYC: Taoyuan City. TNC: Tainan City. KC: Kaohsiung City. NC: Nantou City.

flood scenes obtained through this camera network are shown in Fig. 2(b). These monitoring scenes involve a highly complex environmental background and multiple different view perspectives. Moreover, the resolution and encoding format of the streams are varied because of the different specifications of the camera networks. Table II lists the stream sources and the corresponding resolution and encoding format. Each camera information of IoC including streaming URL, geographic coordinates, and road section, etc., is stored in JSON format (please refer to the format of a JSON object in the Appendix).

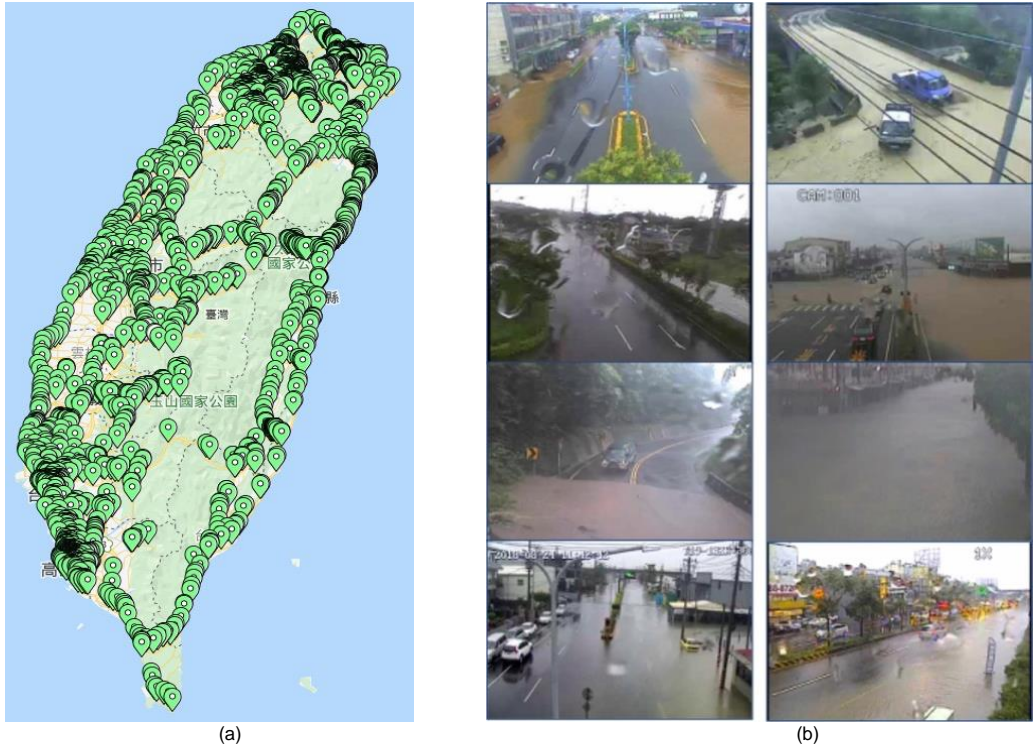

(a) (b)

**FIGURE 2.** Island-wide internet of cameras (IoC). (a) Distribution of the cameras used in this study. The locations of the cameras correspond to the human activity density. (b) Sample flood scene obtained from the IoC.

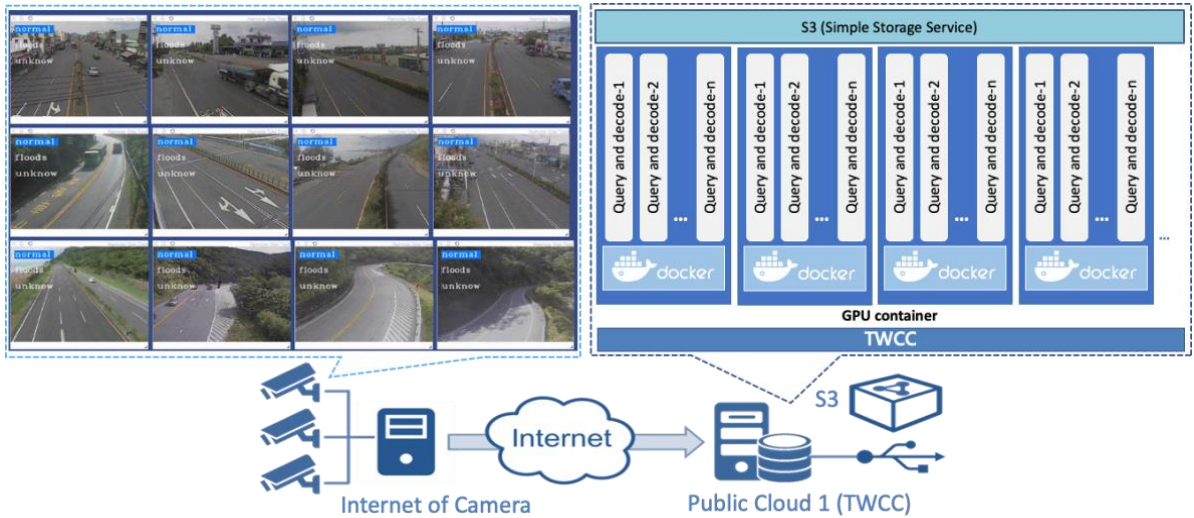

**FIGURE 3.** Image acquisition from the camera network. In the TWCC public cloud, each docker block represented a GPU container involving multiple processes to capture the video stream. Next, each process accessed the IoC video stream through the internet to obtain a real-time scene. In the current setup, the optimal number of query-and-decode processes was 256.

### B. Collection and Encoding of the Image from the Scene Sources

The module described in this section obtained real-time surveillance images from the camera networks. Approximately 2379 CCTV real-time images were captured and encoded to the JPEG format synchronously on the cloud platform. Considering the numerous cameras to be processed, a multiprocessing technique was used to capture streams in a parallel manner. In particular, an efficient approach to acquire large numbers of images is to use multiple processes to execute the task calls asynchronously, in which each process is responsible for a URL of the camera source to capture, decode, and encode a frame from the video stream.

Considering the uncertainties in the network and absence of the source, each process is preset to capture the video stream for 15 s. Subsequently, the process is automatically stopped, and the memory is released. In the true event trial, the collect and decode module processed a camera network in less than 1 min. In other words, the time interval between the individual scenes can be less than one minute, which corresponds to near-real-time simultaneous collection. After the complete stream capture, the flooding sensing module in the next stage performed the scene recognition through the complete batch of the images.

At this stage, the container computing service (CCS) of the TWCC was used as the public cloud. In particular, TWCC has 2,016 NVIDIA Tesla V100 32 GB GPUs. Each CCS can use up to eight V100 GPU accelerators, with 32 CPU cores and 720 GB of memory. Fig. 3 shows the IoC scene collection workflow and multiprocessing layout in the TWCC. Each docker block represents a container in which multiple processes are implemented to manage massive video streaming.

In this study, all the modules were deployed on a container with a minimum set of one GPU, four CPUs, and 90 GB memory. According to previous experiments, the optimal number of processes was 256. Under this setting, the streaming capture could be completed in one minute, and the complete end-to-end workflow involving a round of image capturing of the IoC and providing the final recognition information required five minutes.

For the considered IoC scale, one CCS was sufficient (processing for 2.4k cameras completed in five minutes). However, the system can be efficiently scaled up to increase the number of CCSs to account for an increased number of cameras. Because image files corresponded to the largest demand for the data storage, the simple storage service (S3) cloud data repository was used as a high-performance file-sharing space, and the decoded images could be automatically input and output seamlessly among the cloud containers. The shared S3 repository was created and maintained in the TWCC for the decoded images, and the module config files and outcome JSON files were stored in the local hyper-file system in each container to ensure secure and efficient data access.

### C. Flood Sensing

Flood sensing involves efficiently screening thousands of camera sources to promptly determine whether a flood event is occurring in the scene. In this module, a CNN-based image classifier was used to screen all the collected scenes. If the input scene was classified as that involving waterlogging, it was transferred to the next stage for flood event mapping (III.D) and subsequently for waterlogging notification (III.E).

#### 1) URBAN FLOOD DATASET

To prepare the target domain data to train the neural network, we divided 19,622 images into three labels, namely, normal

TABLE III
ARCHITECTURE OF MOBILENET. EACH LAYER REMAINS UNCHANGED EXPECT FOR THE TOP LAYER (CLASSIFIER), WHICH IS MODIFIED THROUGH THE NEW CLASSES FOR THE TARGET DOMAIN.

| Type / Stride | Filter Shape | Input Size |
|---|---|---|
| Conv / s2 | $3 \times 3 \times 3 \times 32$ | $224 \times 224 \times 3$ |
| Conv dw / s1 | $3 \times 3 \times 32$ dw | $112 \times 112 \times 32$ |
| Conv / s1 | $1 \times 1 \times 32 \times 64$ | $112 \times 112 \times 32$ |
| Conv dw / s2 | $3 \times 3 \times 64$ dw | $112 \times 112 \times 64$ |
| Conv / s1 | $1 \times 1 \times 64 \times 128$ | $56 \times 56 \times 64$ |
| Conv dw / s1 | $3 \times 3 \times 128$ dw | $56 \times 56 \times 128$ |
| Conv / s1 | $1 \times 1 \times 128 \times 128$ | $56 \times 56 \times 128$ |
| Conv dw / s2 | $3 \times 3 \times 128$ dw | $56 \times 56 \times 128$ |
| Conv / s1 | $1 \times 1 \times 128 \times 256$ | $28 \times 28 \times 128$ |
| Conv dw / s1 | $3 \times 3 \times 256$ dw | $28 \times 28 \times 256$ |
| Conv / s1 | $1 \times 1 \times 256 \times 256$ | $28 \times 28 \times 256$ |
| Conv dw / s2 | $3 \times 3 \times 256$ dw | $28 \times 28 \times 256$ |
| Conv / s1 | $1 \times 1 \times 256 \times 512$ | $14 \times 14 \times 256$ |
| 5× Conv dw / s1 | $3 \times 3 \times 512$ dw | $14 \times 14 \times 512$ |
| 5× Conv / s1 | $1 \times 1 \times 512 \times 512$ | $14 \times 14 \times 512$ |
| Conv dw / s2 | $3 \times 3 \times 512$ dw | $14 \times 14 \times 512$ |
| Conv / s1 | $1 \times 1 \times 512 \times 1024$ | $7 \times 7 \times 512$ |
| Conv dw / s2 | $3 \times 3 \times 1024$ dw | $7 \times 7 \times 1024$ |
| Conv / s1 | $1 \times 1 \times 1024 \times 1024$ | $7 \times 7 \times 1024$ |
| Avg Pool / s1 | Pool $7 \times 7$ | $7 \times 7 \times 1024$ |
| FC / s1 | $1024 \times 1000$ | $1 \times 1 \times 1024$ |
| Softmax / s1 | Classifier | $1 \times 1 \times 3$ |

(15295 images), floods (3689 images), and unknown (638 images). The images were recorded in all the locations of the IoC during a real event. The validation set involved 20% (3924 images) of the training set and was not included in the training process. The performance of the trained neural network was evaluated through the validation set.

#### 2) TRANSFER LEARNING FOR FLOOD INSTANCES

In practice, training a complete CNN from the initial state requires a sufficiently large target dataset. However, most such sets for a particular application are inadequate. In this study, the data set contained scenes from specific normal, waterlogging, and unknown situations under various weather and light conditions. The complete CNN was not trained from the initial state, and a pre-trained model on the extremely large ImageNet dataset was employed, which contained 1.4 million images and 1000 categories [20]. This pre-trained model did not include the final fully connected layer, and a fixed feature extractor was used for the flood scene dataset. After fine-tuning, the model was retrained using the new Softmax classifier through the new target dataset. This process corresponded to transfer learning [44].

The key concept of transfer learning is to retrain the model from the existing weights to new target classes [45, 46]. In this work, the source model was the MobileNet [47] network pre-trained on the ImageNet dataset. The choice of neural network architecture is based on accuracy density (representing how efficiently the model uses its parameters) and the need for super real-time performance (0.61~3.34 ms per image for MobileNet) [48]. MobileNet represents an efficient model for mobile and embedded vision applications and is based on a streamlined architecture that employs depth-wise separable convolutions to build lightweight deep neural networks. Therefore, this model is suitable for applications that require the rapid screening of a large number of input images. The source model retrained the final layer for the new target dataset, whereas all the other layers remained unchanged. Table III presents the MobileNet neural network, which maintained the weight of the feature extractor and only employed the new images from the target domain to retrain the classifier. To automatically recognize scenes from the IoC, the retrained neural network model was used as the flood scene classifier.

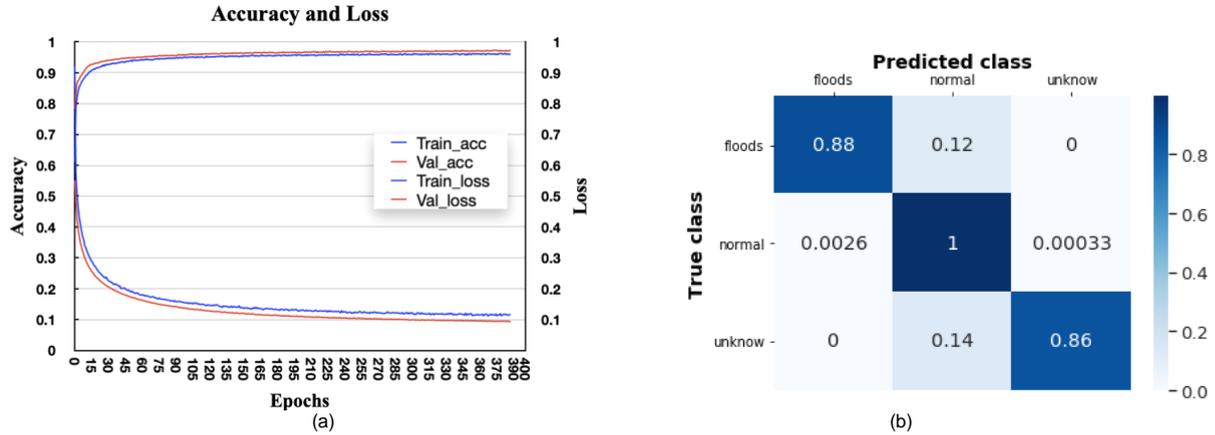

**FIGURE 4.** (a) Training log with accuracy and loss for the training and evaluation dataset, and (b) normalized confusion matrix for each class. Note that the predicted accuracy of the normal class (central) was 0.99707 (automatically rounded to 1).

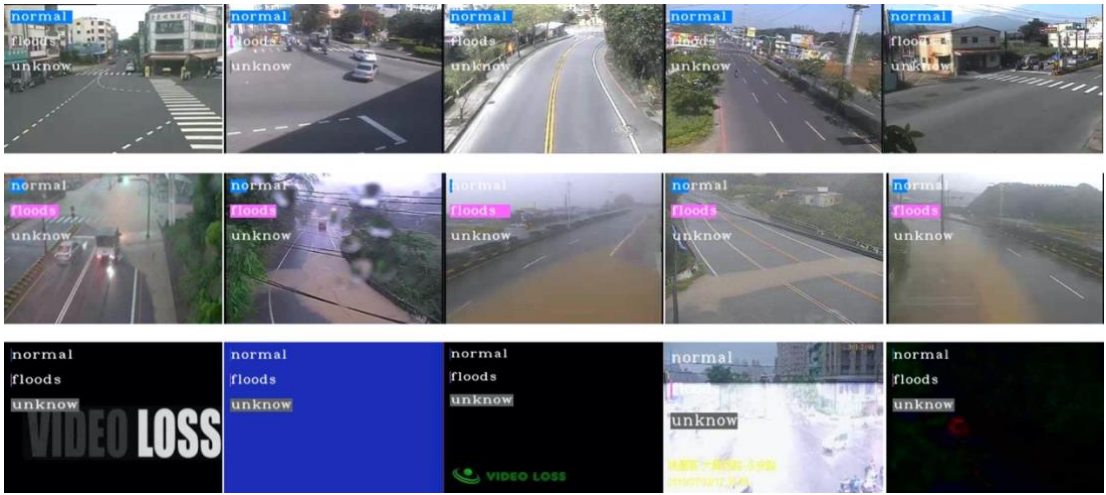

**FIGURE 5.** Example of the scene recognition result from the model. The top, middle, and bottom rows show the results for the normal, flood, and unknown scenes, respectively. The label bars represent the confidence scores for each class (in percentage points).

This MobileNet was implemented by using Keras with a TensorFlow backend and is based on pre-trained weights of the ImageNet dataset. All images were automatically resized to 224×224×3 pixels by the input layer when being fed to the model. The model training was performed on one NVIDIA Tesla V100 with hyperparameters as follows: trained for 500 epochs using the Adam optimizer [49] with an initial learning rate of 1e-2; the learning rate was decreased by a factor of two if the validation loss had no improvement for 5 epochs (patience = 5 epochs) until the lower bound of the learning rate of 1e-8; the batch size was 32; and the early stopping callback with the patience of 10 epochs. The final model weights were taken from the training epoch with the best value of the monitored validation loss.

After the training process, each image of the training dataset was fed to the network to derive the predictions. Next, the predictions were compared to the actual labels to update the weights of the final layer through the back-propagation process. Finally, this model was retrained with the flood dataset to distinguish among different scenes. Fig. 4(a) shows the results for the training and validation set along with the confusion matrix. The model trained for 388 epochs, and the training and validation accuracies were 96% and 97%, respectively. The training process was terminated via an early stopping scheme when the validation loss did not decrease in the last 10 epochs. Fig. 4(b) shows the confusion matrix for the classification results of the trained model over the validation dataset.

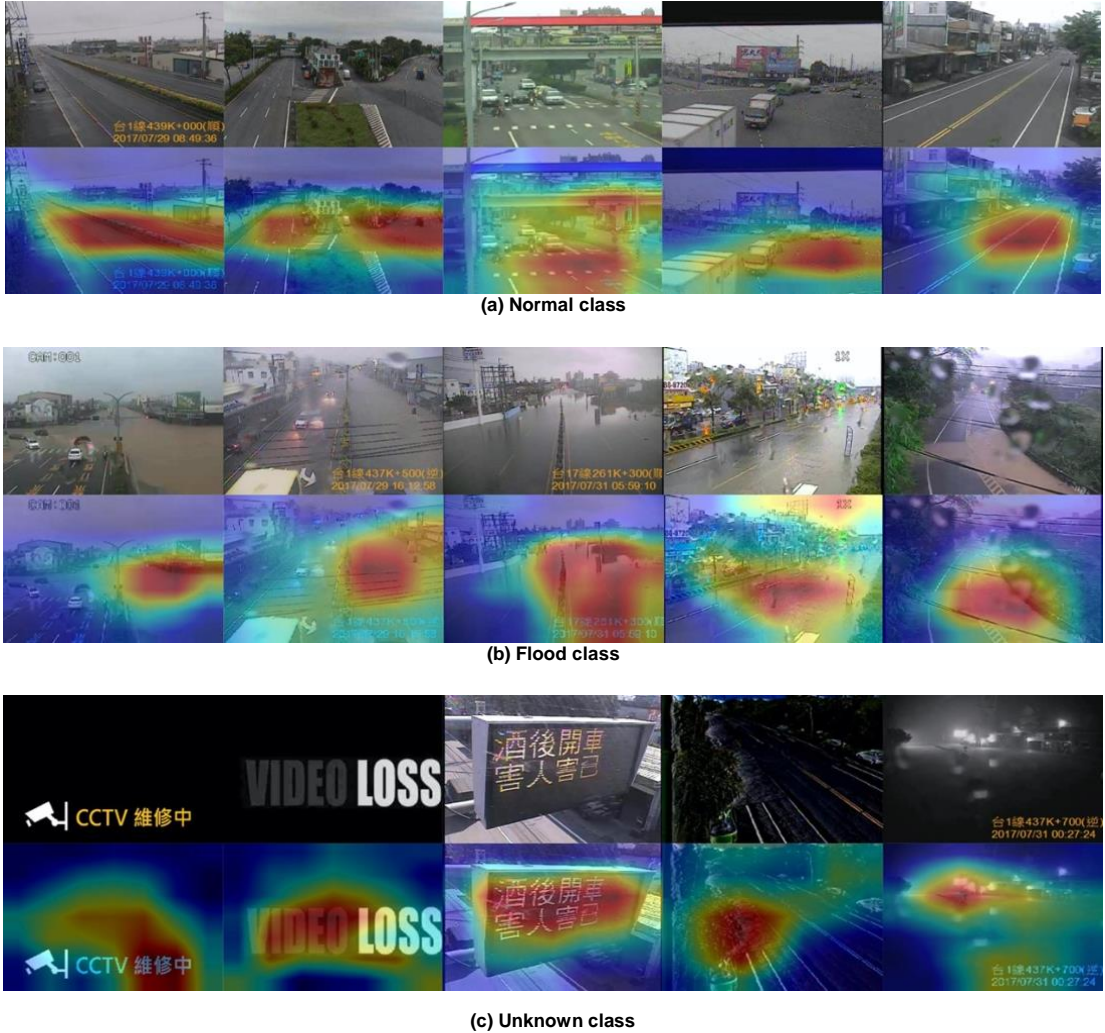

**FIGURE 6.** Example of the decision attention of the trained model. The upper and lower rows show the input scene and Grad-CAM heatmap for the normal, flood, and unknown classes, respectively.

The trained model was used to identify the IoC scenes. Several representative results are shown in Fig. 5, in which the top, middle, and bottom rows correspond to the normal, flood, and unknown scenes, respectively. The unknown scenes include the scenes with signal loss, harsh noise, damaged video streaming, and incorrect field of view. To illustrate the proportion of each class feature, the confidence value is displayed on the upper left of the figure as a bar graph. The length of the bar graph represents the confidence value in percentage points. The background features of the flood scene are similar to those of the normal scene; therefore, the confidence value is partially biased toward the normal class. However, the class of the scene can still be distinguished through the magnitude of the confidence value.

3) VISUAL EXPLANATION OF THE MODEL DECISIONS
Usually, the accuracy and loss of the training and validation can adequately represent the model (trained networks) performance for a specific dataset. The accuracy/loss metrics are simple and can directly represent the model's ability to identify scenes with different categories. However, the decision behavior of the trained model is implicit, and a human cannot easily understand the mechanism adopted by a neural network model to attain the outcome. To clarify the behavior of the model inference, the gradient-weighted class activation mapping (Grad-CAM) method [50, 51] was used to demonstrate the role of feature judgment, that is, the process in which model seeks the decision basis. Fig. 6 presents the Grad-CAM heatmap for each class example. The examples demonstrate that in most cases, the model focuses more on the related regions from which the scene class can be inferred. Furthermore, to confirm and explain how to learn the right region for a specific class's features with the trained model, Fig. 6(a) shows that the "Normal class" focused on the dry area of the road, Fig. 6(b) shows that the "Flood class" focused more attention on the waterlogged areas, and Fig. 6(c) shows that the "Unknown class" focused on unrelated background objects. Moreover, the unknown class contained images other than normal and flood class. Most of them were low-quality images with much noise, unstable or interrupted streaming signals, and the wrong field of view. Therefore, Fig. 6(c) shows a few abnormal samples outside the normal and flood class. The first two are signal sources interruption, the third is the wrong field of view, and the last two samples are poor image quality. These images are likely to influence the performance of normal/flood judgment, hence they are used as the third category for discrimination.

### D. PLUGGABLE WATER LEVEL SENSING MODULES

As presented in the introduction, the current "ruler-free" water level estimation methods have their own limitations (proper shooting angle, field of view, distance between camera and reference object, reference object existing in the scene, scale calibration of reference object, etc.) and various uncertainties could affect the availability of water level estimation (image quality, visibility of reference object, actual size of reference object, etc.). In the proposed system, these water level estimation methods are only used in a small number of specially selected monitoring locations, owing to the above-mentioned reasons. In order to give some context to the "ruler-free" water level sensing methods, we present a brief overview of our implementation, which is based on object detection and segmentation networks.

#### 1) UNIVERSAL REFERENCE

In large-scale visual sensing, the cost may too high to set up the water level ruler for most of the cameras [17, 18]. Furthermore, it is also difficult to find a consistent and common object as a reference ruler in all scenes. Therefore, these conditions limit the availability of vision-based water level estimation.

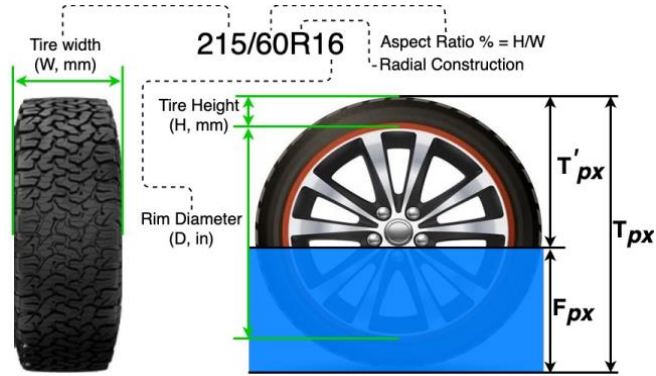

**FIGURE 7. Tire marking and size, and the relationship to flood height. This tire marking example of 215/60/R16 represents the width, aspect ratio, radial construction, and rim diameter. Further, to grade the water level, the scale relationship between wheel ($T_{px}$) and flood ($F_{px}$) is defined as $F_h$.**

Considering a reference object that is available in most urban scenarios (a universal reference), and to avoid repeated measurement and calibration of the various reference objects, this study uses the wheel of a sedan as the reference for the fast flood status grading and estimation of the water level. The main size mark definition of a tire is shown in Fig. 7. The aspect ratio (H/W) defines the relationship between tire width and sidewall height in percentage (excluding rim diameter, D). The higher the aspect ratio, the higher the height of the tire. As shown in (1), the wheel diameter can be calculated from the size marking. We need to know the rim diameter and the tire height in order to calculate the diameter of the wheel. The rim diameter is already given by size marking. The tire height (tire sidewall) can be calculated from the aspect ratio. For example, a tire with the size marking 215/60R16 has the following:

1) tire width ($W_{tire}$) of 215 mm,
2) aspect ratio ($A_{ratio}$) of 60 %,
3) radial layers (R),
4) rim diameter (D) of 16 inch (40.64 cm),
5) tire height (H) of 129 mm (215 mm × 60%).

The diameter of the wheel ($D_{wheel}$ = 66.44 cm) as follows:

$$D_{wheel} = D + (W_{tire} \times A_{ratio} \times 2) \quad (1)$$

where D is the rim diameter in inches, $W_{tire}$ is tire width in centimeter, and $A_{ratio}$ is the aspect ratio of the tire in percentage (%).

The geometric relationship between the water level and the wheel is also shown in Fig. 7, where

1) $F_{px}$ is the depth of flooding in pixel,
2) $T_{px}$ is the entire height of the wheel in pixel,
3) $T'_{px}$ is the height of the non-submerged part of the wheel in pixel.

Thus, $F_h$ is the water level in unit of wheel range from zero to one, and can be obtained from

$$F_h = F_{px}/T_{px} = 1 - T'_{px}/T_{px} \quad (2)$$

where $F_L$ is the water level in centimeter, and can be obtained from

$$F_L = F_h \times D_{wheel} \quad (4)$$

The goal is to take into account the universality of the car itself and, in general, the fact that the common wheel size of a car is about 14 inches to 18 inches, with a maximum possible wheel diameter range of about 55 to 75 cm. Fig. 8 shows the possible wheel diameter range under different tire specifications. According to Taiwan's local disaster compensation regulations, a flooding level greater than 50 cm is the basis for disaster compensation[1]; thus, half the height of the wheel is used as the preliminary classification rule for rapid warning. Therefore, the tires are especially suitable as a reference for the fast grading of floods.

---

[1] Disaster Prevention and Protection Act (https://law.moj.gov.tw)

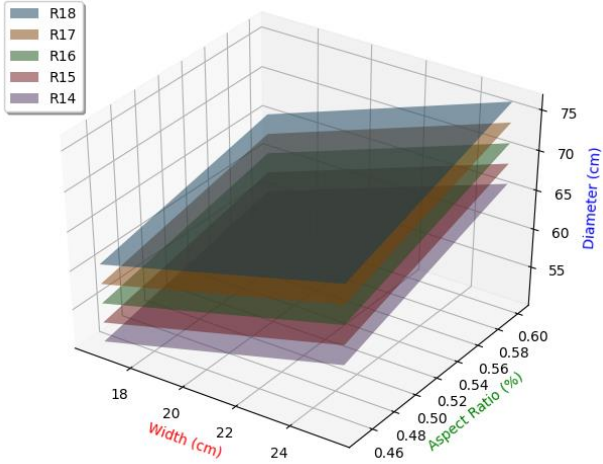

FIGURE 8. Wheel diameter range according to size parameters. For simplicity, to determine the flooding level for alarm, this study assumed a 215/60R16 ($D_{wheel}$ = 66.44 cm) tire for the sedan type vehicle.

#### 2) FLOOD GRADING AND FLOOD LEVEL ESTIMATION BY OBJECT DETECTION

This module uses the YOLO [52] network for rapid grading of the flood level and is treated as a double check module for the previous output of the flood detection network. The concept of grading is to use the tires of a sedan as the scale reference to estimate the flood level indirectly.

Flood grading utilizes the vehicle detection model and the wheel detection model. The vehicle detection model uses the pre-trained model from the original paper [52], and in the detection results objects other than the vehicle are filtered out. Then, the wheel detection model detects the wheels in the bounding box of the detected vehicle. Note that the reason for not directly detecting the wheels is to avoid the reduction of the detection efficiency when the wheels are submerged more than 2/3. Furthermore, using the bounding box of the detected vehicle limits the relative position of the wheels to where they can be used as a basis for exception judgment when the flood exceeds the wheels or a flood does not exist. Additionally, only performing wheel detection in a limited bounding box can reduce the amount of calculation and speed up processing performance.

There are only two types of images in the flood grading dataset to train the wheel detection model (also based on the YOLO): flooding above 1/3 of the wheel (84 instances) and no flooding (307 instances). An example of the training set with the bounding box of ground truth is shown in Figs. 9(a) and (b). Consequently, the wheel detection model outputs the direct grading of a flood that indicates whether the flooding is higher than 1/3 of the wheel or not. Figs. 9(c) and (d) show the results of the flood grading from the wheel detection model. Note that the confidence value shown in Figs. 9(c) and (d) is not used as a basis for determining the flooding height. It is only used for grading scenes as flood or flood-free.

The tire detection result also can be used to calculate the flooding level in centimeters ($F_L$). As shown in Fig. 7, the flooding level in pixel ($F_{px}$) is simply obtained by the upper bound of water inside the bounding box of the detected wheel, and $T_{px}$ is the height of the bounding box in pixels. Therefore, the $F_h$ is the flood level in the unit of the wheel obtained from (2). The value of $F_h$ is between zero to one and represents the proportion of the water level to the wheel height. While the flood level in the unit ($F_h$) is estimated from the wheel detector, the flood level ($F_L$) in centimeter is obtained by multiplying by the wheel diameter $D_{wheel}$, as in (3). However, the wheel diameter varies with the specifications and will affect the final $F_L$ value. Therefore, this module reserves $D_{wheel}$ as an input-able parameter to meet the present conditions of different regions or countries. The most common tire size in Taiwan was chosen as the basis of the calculation. In the present results, the default is set to 215/60R16 ($D_{wheel}$ = 66.44 cm) tire for the sedan type vehicles (as shown in Fig. 8). The experimental results of the flood level in units ($F_h$) are shown in Figs. 9(e) and (f).

#### 3) FLOOD LEVEL ESTIMATION BY SEMANTIC SEGMENTATION

As explained above, we used the height difference between the tire and the upper bound of water to estimate the flood level from the wheel detection network. In this module, we use a segmentation network, Mask R-CNN [53], to detect the wheel location at the pixel level rather than the bounding box. Previous research showed that using the segmentation network to detect the wheel and calculate the flood depth is an effective approach [29]. For more information about the water level estimation with the segmentation network, please refer to the original paper [29].

The advantage of the wheel segmentation network is that both flooded and flood-free instances can be used to train the network to learn the wheel features. Generally, the presence of non-flooded images is more frequent and easy to collect rather than the flooded images. The slight drawback is that the time cost increases when annotating the ground truth for the wheel at the pixel level. The training dataset is the same as the previous object detection approach but with additional mask annotation to train Mask R-CNN. The results of wheel segmentation from the trained network are shown in Fig. 10(a). Moreover, Figs. 10(b) and (c) show that water level estimation is easily affected by the image quality and water waves. Finally, Figs. 10(d) to (g) show the base pixels from the segmented mask used for water level estimation.

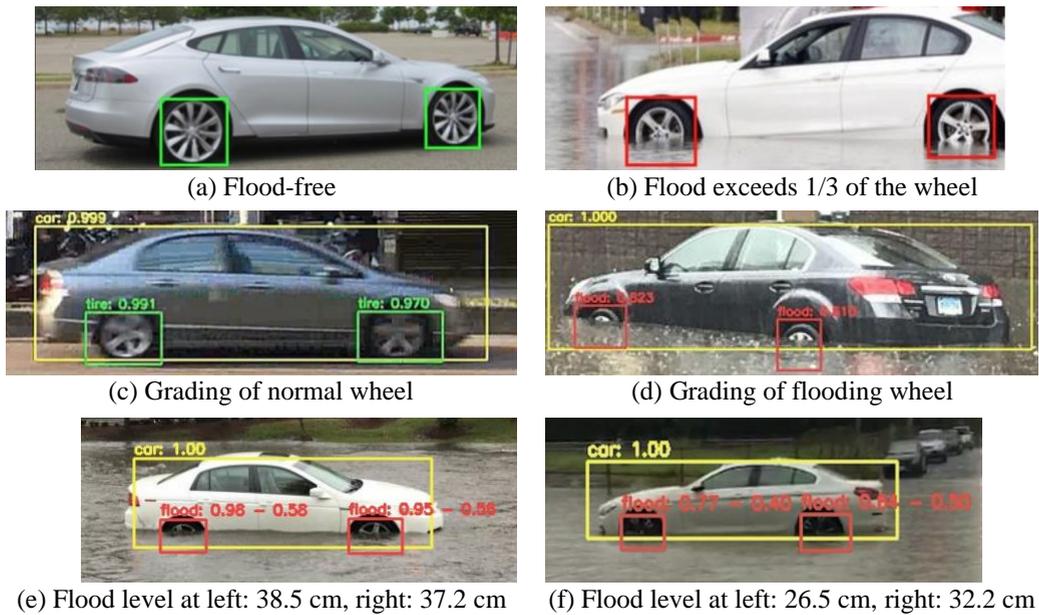

FIGURE 9. Result of flood grading by wheel reference. The training set contains two identical types of (a) flood-free and (b) flood wheel instances. (c) Detection result of a normal scene. (d) Detection result of a waterlogged scene. The labels and bounding boxes represent the normal tire (green) and flooded tire (red) and its confidence scores respectively. (e-f) Additional flood levels in unit ($F_h$) based on the wheel, and the flood level in centimeter ($F_L$) are (e) left: 38.5 cm, right: 37.2 cm and (f) left: 26.5 cm, right: 32.2 cm based on 215/60R16 ($D_{wheel}$ = 66.44 cm). Note that the annotation in (e-f) is "flood: (confidence score) – ($F_h$)."

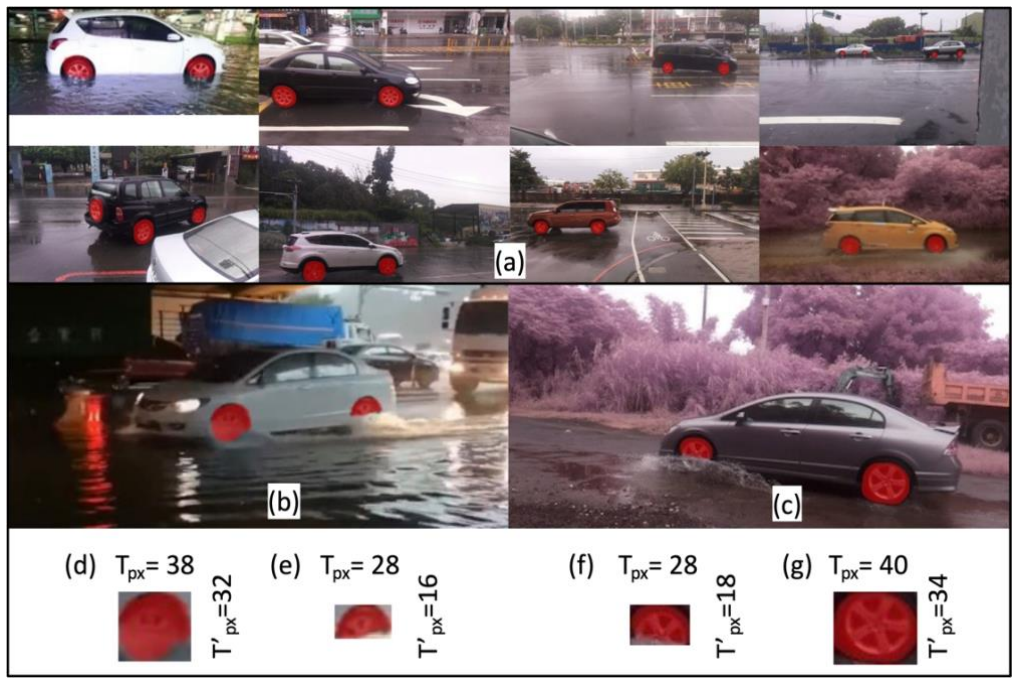

FIGURE 10. Results of wheel segmentation and water level estimation. (a) Wheel segmentation results from various relative distances and perspectives to the objects. (b-c) Case of water wave occluding a part of the wheel as the car passes through the water at a relatively high speed. (d-g) Water level estimation of (b-c) according to (2); then, $F_L = 1 - (T'_{px}/T_{px})*D_{wheel}$, when $D_{wheel}$ = 66.44 cm, resulting in (d) $F_L$=10.5 cm, (e) $F_L$ = 28.4 cm, (f) $F_L$ = 23.7 cm, (g) $F_L$ = 9.9 cm.

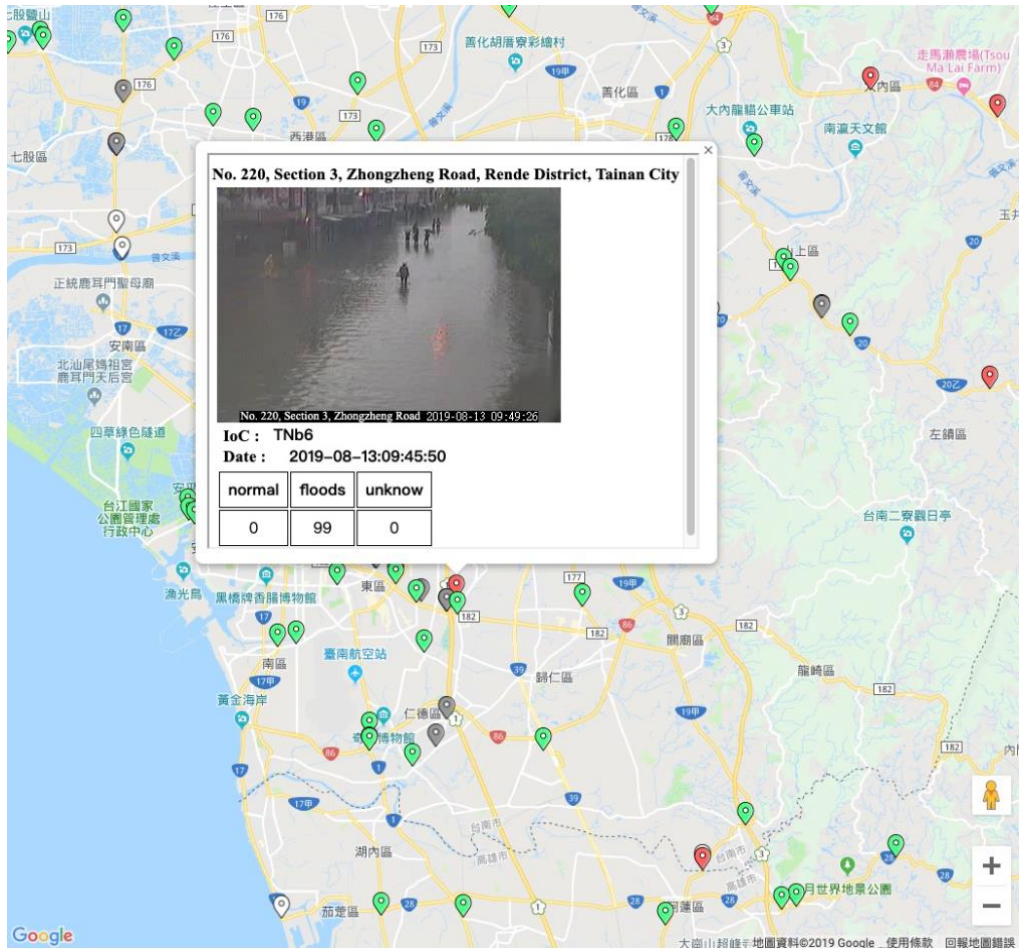

**FIGURE 11.** Real-time in situ video stream viewing at the GIS map (same as that shown in Fig. 2(a), albeit zoomed-in). Each location marker indicates an in situ camera marked in red (flood), green (normal), gray (unknown), or white (no video). Clicking each marker led to the popping out of a sub-window that displayed the real-time video streaming, description of the street address, camera ID, timestamp of observation, and probabilities for each class from the classification model (see Section III.C).

### E. Geo-Referenced Flood-Mapping

The resulting flooded location from the flood sensing phase was updated in the web-based GIS map. This map was public to the citizens and could update the flood information in intervals of 5 min or longer according to the requirements. In periods of rainstorm advisory, such as the periods of weather warning from the Center Weather Bureau, the system could achieve minimal time latency in 5 min. In other situations, the interval could be increased to an hour or more to reduce the computing and storage resource.

Fig. 11 shows the detected flood scene from the IoC, with each possible flooded location marked as a red point to enhance the identifiability. The GIS map provided the island-wide mapping of each scene location of the IoC, and the pop-sub-window relayed the real-time video stream when the user clicked the marker point. Fig. 11 corresponds to the same map shown in Fig. 2(a), albeit with a zoomed-in view to clarify the details. In particular, Fig. 11 shows a snapshot of the system trial run from the rainstorm event in south Taiwan on August 13, 2019.

Fig. 12 shows a special case for the false alarm of flood detection. Under the same rainstorm event, the system recognized a camera at the farther south location as a flood and marked it in red as an alarm point. However, this scene did not correspond to a flood event; the scene involved a bridge over an overflowing river, with the water submerging the pipes. In this context, the trained network performed an accurate inference for this scene, as the muddy water covered the ground. In such cases or other false alarm cases, the users should refer to the onsite video stream for the final decision-making. In other words, onsite images for reference must be provided when the system broadcasts the flood information to the citizens. The subsequent section describes the notification module that compiled both the visual clue with the flood-mapping information to the authorized users.

### F. Reporting and Notification Agent

In the aforementioned processes, the visual sensing system performed flood recognition and updated the instance information to the flood map. Although the citizens could easily access the flood map through a browser and search and navigate the flood events on the map, the flood information was a passive announcement. A specific group of users may require concise and rapid dissemination of the information regarding the flood development state. Thus, a notification agent was used to provide disaster prevention crews with the latest waterlogging information immediately. These instant infographics were not directly sent to the citizens, but the information could be accessed at the flood-mapping web service page, as described in Section III.E.

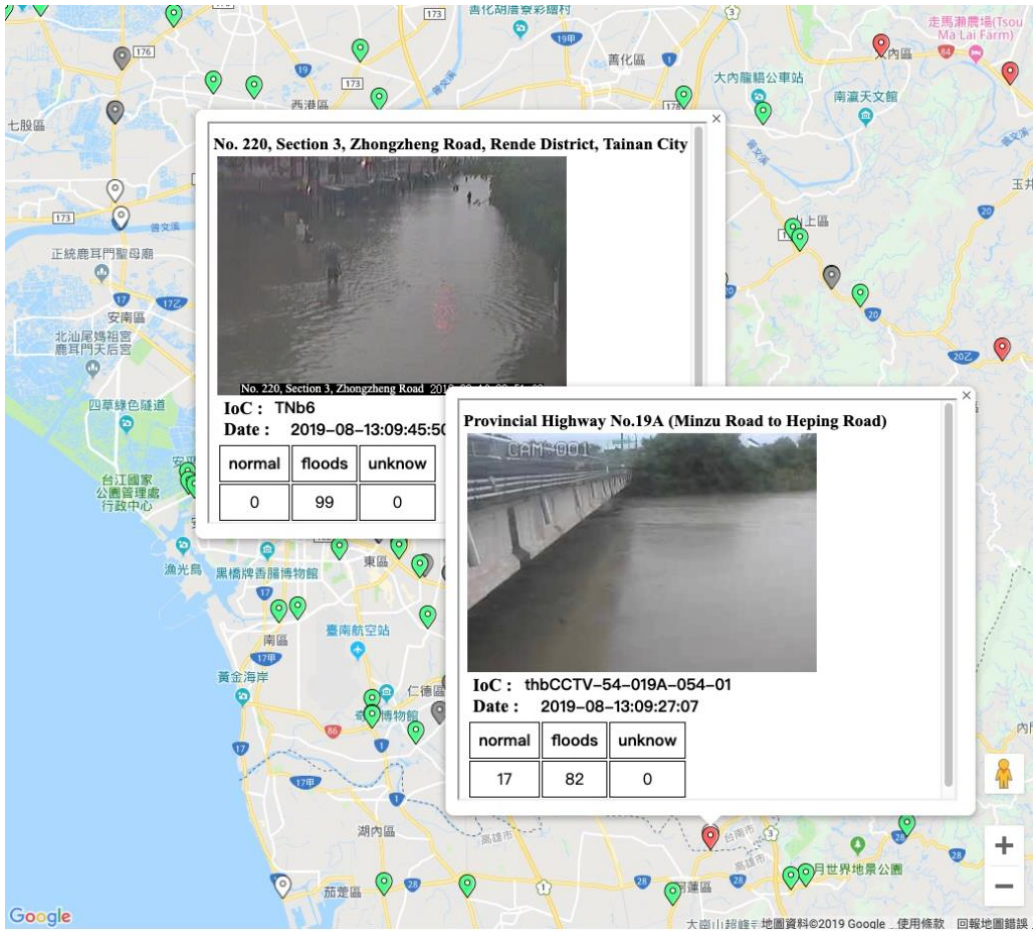

**FIGURE 12.** Unique case of false alarm from the visual sensing system under the same rainstorm event. As shown in the lower-right pop-window, the trained network misjudged the overflowing river as a flood (confidence 82% for a flood). However, this judgment was actually correct in terms of the visual characteristics.

The notification agent used a filtered map that only displayed the flood event and annotated the address and corresponding image of the events over the flood map. Fig. 13 shows the concise report pertaining to the flood situation. This report represented a sub-version of the flood GIS map, which only showed the flood events with each event numbered from north to south. Moreover, this report showed the annotated road section on the left of the map and the appended corresponding scenes on the right side. Through the summary report of the visual sensing, specific users could be informed of the current disaster situation and onsite image information, which could effectively allow the users to promptly respond to the disaster. In addition, this concise map could be used as a flood GIS map in the subpage without the flood scenes.

In the final phase, the agent sent the instant visual report to the LINE group to enable prompt awareness enhancement and response capabilities, along with a URL link of the flood map to provide detailed information. LINE is a freeware app to realize instant communication on electronic devices such as smartphones, tablet computers, and personal computers. LINE users can exchange texts, images, video and audio, and conduct free VoIP conversations and video conferences [54]. LINE is the most popular instant messaging app in Taiwan, with a penetration rate of 86% for internet users aged from 16 to 64 y [55].

Fig. 14 shows the immediate flood report received by a LINE user. In this app, to obtain detailed information, the user can tap the URL link from the bottom of the report to access the web portal of the flood map to check the flood events and corresponding real-time video. Moreover, the user can check the current event situation through the GIS map simply via a browser. Each clickable red flag marker in the map represents an in situ camera that provides real-time video and the corresponding description of the street address, camera ID, timestamp of observation, and probabilities for each class from the classification model (as described in Section III.C). Based on this report, the user can immediately perceive the flooding distribution and realize in situ viewing via a single visual graph that can help the citizens respond promptly to the disaster.

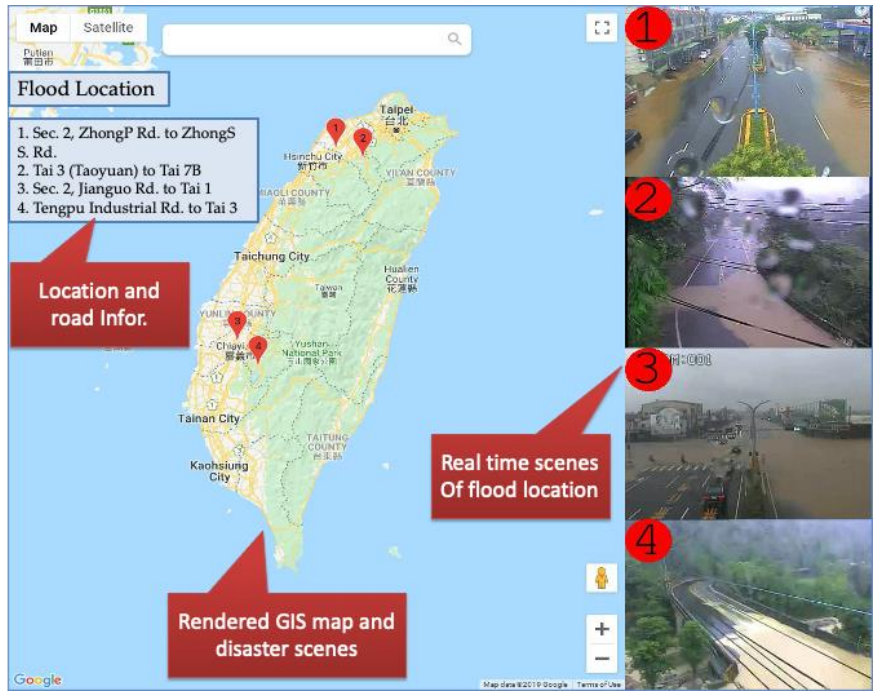

**FIGURE 13.** Summary report of the visual sensing with real events from the test case. The report presents a near-real-time map of the detected flood scenes and incidents in graphical form. The individual events are numbered sequentially from north to south according to the latitude for easy identification. The "flood location" block on the upper left and road name block below it, similar to the live scene on the right, are automatically generated in the summary report.

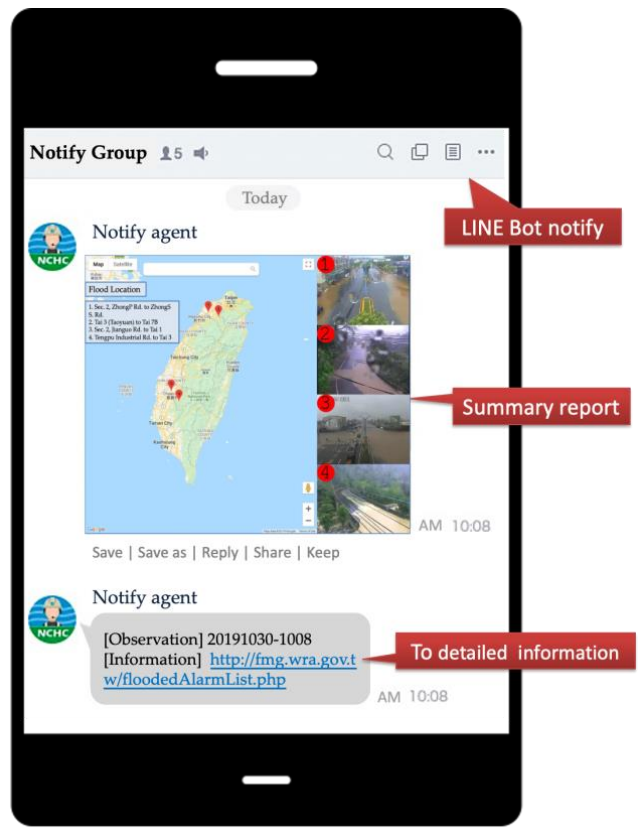

**FIGURE 14.** The LINE bot agent automatically sends the current localization mapping of the waterlogging to the LINE group. This auto notification is for disaster management and research and only sent to specific users. The final summary report of the notification agent for the real event from the test cases is also shown, which corresponds to the immediate rendering of the GIS map and detected scenes.

## IV. DISCUSSION

This research demonstrated a scalable deep sensing workflow to be used during island-wide flood events. The proposed approach is based on public cloud computing resources with deep CNNs to provide end-to-end flood perception services.

The deep sensing system has two main functions: it can efficiently discriminate a large number of scenes from the IoVT/IoC and provide/update the flood geo-referenced map over-the-air. The final decision-making and response are implemented by the users in different roles to avoid the flood sections or formulate an action plan to alleviate the impact of the flood event. The system was applied to flooding events in 2019. The current system can process a 2.4K input from the IoC and output results in five minutes, provide preliminary examinations for disaster prevention units, and assist the citizens in familiarizing themselves with the system. Future work can be aimed at examining the approaches to effectively combine various sensing information points and perform analyses to assist the disaster prevention department in making decision and ensuring the information correctness and implementing rapid corrections in the case of false information.

The neural network developed in this study can still not surpass human accuracy, and a small number of misclassifications may be displayed on the GIS map. Generally, users can directly refer to the images in the GIS map or the summary report and automatically ignore the misclassified locations. However, a feasible direction in future work is the implementation of a user interaction interface based on the GIS map that allows users to rectify the prediction results by the neural network and use this feedback to dynamically update the neural network and GIS map.

Moreover, this system only used a part of the training set to evaluate the recognition accuracy (theoretical recognition performance of the network model) because in this study, the resources required to acquire a considerable amount of data and perform manual annotation for the actual flooding event were limited. During the system operation in 2019, all the collected images were not fully reserved and properly interpreted. Hence, the complete accuracy of the system was not evaluated on the true flood event. In future work, to suitably evaluate the system performance, a consummate flood scene dataset corresponding to long-period collection and variety of weather events should be established.

The ability to obtain fine-grained flood depth estimation in visual images can assist in extracting the flooding information. In particular, in addition to detecting whether local flooding occurs, the current flooding depth can also be estimated. Nevertheless, the actual situation is more complex. In the latest related research, common reference objects such as pedestrians, bicycles, and car tires were used as virtual ruler to estimate the water depth. However, such applications require specific shooting angles and image quality or use crowdsensing images, which limits the distribution density of the sensing, as well as the sensing timeliness and usability. Therefore, identifying a common reference target under most camera conditions and developing a general water depth-sensing method is a worthy research direction.

In terms of the computational performance, the current system used two computing containers belonging to different public cloud providers to process the image recognition and outcome presentation. The advantage of the cloud platform system is that the number of computing containers and internet connection groups can be rapidly adapted to achieve load balance under varied scales. For an IoC with 2.4K video sources, the overall end-to-end processing can be completed in less than five minutes (including the input IoC streams to output the flood notification); therefore, the released notifications and GIS maps are less than five minutes old. As this system adopts the sensor to the cloud mode, all the sensor data are captured and transmitted to the cloud for batch processing. In other words, the system cannot meet the demand for real-time response owing to the operation latency. In this context, the real-time sensing and onsite analysis of large-scale visual sensors and means to realize a balance of the computing performance, power consumption, and low cost are valuable research directions.

We choose CNN in terms of process automation and recognition performance. There are two end-to-end scenarios: system level, from the input of IoC to GIS map, and inner module level, from the CNN model trains/inferences the input images to output detection results, both without any manual feature engineering and pre-processing. The CNN-based approaches provide an end-to-end pipeline that excludes the usual manual feature engineering process within machine learning and other image processing methods. It also eliminates the uncertainties of variable environmental conditions that most often affect the performance of image processing methods. In addition, we chose the neural network with the higher accuracy density to achieve the goal of large-scale recognition in near real-time. Most of the current studies are based on existing neural networks that have been empirically tested with large open datasets. However, comparing different depths, widths, and architecture of CNNs may be interesting in future work. Within the scope of our knowledge, there has not been a similar quantitative study on the neural network for flood recognition.

According to the research pertaining to image recognition tasks, AI models can easily reach or surpass the so-called human-level performance (HLP), which is also the purpose of research and experimentation. However, even if a model surpasses the HLP in terms of the test-set accuracy, the system is not necessarily superior to humans in practical applications in the real-world. In practice, the objective of researchers and disaster management administrators is not limited to surpassing the HLP in terms of the test-set accuracy. The tolerance, latency, bias, performance on rare scenes, and additional factors must be considered when deploying the system. This aspect can be attributed to the extensiveness of the training data and the inference performance of noisy data and rare classes. Therefore, another possible future research direction is to improve the HLP by exploiting the superior performance of AI models. Specifically, AI can facilitate the decision-making of humans instead of completely replacing or outperforming human decision-making.

At present, owing to the lack of long-term flooding data collection and correct labeling in the flood identification domain, AI is lacking in terms of long-term real-world identification performance testing. Therefore, in real-world disaster applications, an acceptable system may exhibit low false negatives and high false positives, because it is more desirable to display a larger number of possible flooding events (regardless of the trueness of the information) than miss an opportunity to identify a flooding event owing to a false negative. The deep sensing system can simplify all the inputs and reduce their number, leaving only a few more possible events. Such a design can simplify the information extraction and decision-making process.

This research involved the development of an end-to-end flood sensing and response system based on IoVT/IoC on an island-wide scale. This research helped examine the various scenarios that may be encountered in the practical operation of large-scale cyber-physical systems, and the observations can serve as an empirical reference to develop visual sensing systems.

## V. CONCLUSIONS

A scalable deep sensing workflow with real flood events on an island-wide scale was developed. The proposed workflow involves a cognitive sensing system for large-scale waterlogging perception based on deep learning models, information and communications technology, and cloud resources. The system has enabled plug-in functions, allowing flexible changes to the visual sensing module. Therefore, the end-to-end system architecture and key modules may be utilized in other types of disaster events for damage/impact assessment, such as wildfires, land slippage, urban fire, and explosions. From a practical viewpoint, the proposed system can offer an efficient and cost-effective alternative to provide rapid waterlogging screening reports on-the-air to facilitate deep remote sensing.

## APPENDIX

## JSON OBJECT OF IoC

```
IoC =
{
  "DGH":
  [
    {
      "tvid": "thbCCTV-12-0090-037-01",
      "Longitude": 121.70156,
      "Latitude": 24.93671,
      "roadsection": "Provincial Highway 9 (Sec. 8, Beiyi Rd.)",
      "url": "http://11.22.33.44/T9-1K+150"
      ...
    },
    {
      "tvid": "thbCCTV-11-0022-044-05",
      "Longitude": 121.62588,
      "Latitude": 25.26965,
      "roadsection": "Provincial Highway 2 (Zhongzheng East Rd.)",
      "url": "http://11.22.33.44:/T2-3K+750",
      ...
    },
    ...
  ],
  "NTPC":
  [
    {
      "tvid": "NTPC-C000001",
      "Longitude": 121.4699,
      "Latitude": 25.0203,
      "roadsection": "Minsheng Rd., Wanban Rd., Banqiao District",
      "url": "http://77.66.55.44/Media/Streaming?deviceid=1"
      ...
    },
    ...
  ],
  ...
  "NC":
  [
    ...
  ]
}
```

## ACKNOWLEDGMENT

We are indebted to Water Resources Agency (WRA) and Directorate General of Highways (DGH) in Taiwan for their video streaming. And we are grateful for the support of National Center for High-performance Computing (NCHC) and Taiwan Computing Cloud (TWCC), and advice from Visual Sensing Lab (VSL).

## REFERENCES

[1] S. Kelly *et al.*, "Impacts of a record-breaking storm on physical and biogeochemical regimes along a catchment-to-coast continuum," *PLoS One,* vol. 15, no. 7, p. e0235963, 2020, doi: 10.1371/journal.pone.0235963.
[2] C. Wolff, T. Nikoletopoulos, J. Hinkel, and A. T. Vafeidis, "Future urban development exacerbates coastal exposure in the

Mediterranean," *Scientific Reports,* vol. 10, no. 1, p. 14420, 2020/09/02 2020, doi: 10.1038/s41598-020-70928-9.

[3] S. Van Ackere, J. Verbeurgt, L. Sloover, S. Gautama, A. De Wulf, and P. De Maeyer, "A Review of the Internet of Floods: Near Real-Time Detection of a Flood Event and Its Impact," *Water,* vol. 11, no. 11, Nov 2019, Art no. 2275, doi: 10.3390/w11112275.

[4] S. Puttinaovarat and P. Horkaew, "Flood Forecasting System Based on Integrated Big and Crowdsource Data by Using Machine Learning Techniques," *IEEE Access,* vol. 8, pp. 5885-5905, 2020, doi: 10.1109/ACCESS.2019.2963819.

[5] U. Iqbal, P. Perez, W. Li, and J. Barthelemy, "How computer vision can facilitate flood management: A systematic review," *International Journal of Disaster Risk Reduction,* vol. 53, p. 102030, 2021/02/01/ 2021, doi: https://doi.org/10.1016/j.ijdrr.2020.102030.

[6] E. Bousquet, A. Mialon, N. Rodriguez-Fernandez, C. Prigent, F. H. Wagner, and Y. H. Kerr, "Influence of surface water variations on VOD and biomass estimates from passive microwave sensors," *Remote Sensing of Environment,* vol. 257, p. 112345, 2021/05/01/ 2021, doi: https://doi.org/10.1016/j.rse.2021.112345.

[7] A. Guo, J. Chang, Y. Wang, Q. Huang, and Y. Li, "Uncertainty quantification and propagation in bivariate design flood estimation using a Bayesian information-theoretic approach," *Journal of Hydrology,* vol. 584, p. 124677, 2020/05/01/ 2020, doi: https://doi.org/10.1016/j.jhydrol.2020.124677.

[8] M. Avand, H. Moradi, and M. R. lasboyee, "Using machine learning models, remote sensing, and GIS to investigate the effects of changing climates and land uses on flood probability," *Journal of Hydrology,* p. 125663, 2020/10/27/ 2020, doi: https://doi.org/10.1016/j.jhydrol.2020.125663.

[9] I. Chawla, L. Karthikeyan, and A. K. Mishra, "A review of remote sensing applications for water security: Quantity, quality, and extremes," *Journal of Hydrology,* vol. 585, p. 124826, 2020/06/01/ 2020, doi: https://doi.org/10.1016/j.jhydrol.2020.124826.

[10] B. Dixon, R. Johns, and A. Fernandez, "The role of crowdsourced data, participatory decision-making and mapping of flood related events," *Applied Geography,* p. 102393, 2021/03/10/ 2021, doi: https://doi.org/10.1016/j.apgeog.2021.102393.

[11] S. W. Lo, J. H. Wu, F. P. Lin, and C. H. Hsu, "Cyber surveillance for flood disasters," *Sensors,* vol. 15, no. 2, pp. 2369-87, Jan 22 2015, doi: 10.3390/s150202369.

[12] S. W. Lo, J. H. Wu, F. P. Lin, and C. H. Hsu, "Visual Sensing for Urban Flood Monitoring," *Sensors,* vol. 15, no. 8, pp. 20006-29, Aug 14 2015, doi: 10.3390/s150820006.

[13] D. G. Costa, "Visual Sensors Hardware Platforms: A Review," *Ieee Sensors Journal,* vol. 20, no. 8, pp. 4025-4033, Apr 2020, doi: 10.1109/jsen.2019.2952447.

[14] S. Jiang, V. Babovic, Y. Zheng, and J. Xiong, "Advancing Opportunistic Sensing in Hydrology: A Novel Approach to Measuring Rainfall With Ordinary Surveillance Cameras," *Water Resources Research,* vol. 55, no. 4, pp. 3004-3027, 2019, doi: 10.1029/2018wr024480.

[15] Y. H. Lou *et al.*, "Front-End Smart Visual Sensing and Back-End Intelligent Analysis: A Unified Infrastructure for Economizing the Visual System of City Brain," *Ieee Journal on Selected Areas in Communications,* vol. 37, no. 7, pp. 1489-1503, Jul 2019, doi: 10.1109/jsac.2019.2916488.

[16] P. K. Bhola, B. B. Nair, J. Leandro, S. N. Rao, and M. Disse, "Flood inundation forecasts using validation data generated with the assistance of computer vision," *Journal of Hydroinformatics,* vol. 21, no. 2, pp. 240-256, 2018, doi: 10.2166/hydro.2018.044.

[17] Z. Zhang, Y. Zhou, H. Liu, L. Zhang, and H. Wang, "Visual Measurement of Water Level under Complex Illumination Conditions," *Sensors (Basel),* vol. 19, no. 19, p. 4141, Sep 24 2019, doi: 10.3390/s19194141.

[18] G. Chen *et al.*, "Method on water level ruler reading recognition based on image processing," *Signal, Image and Video Processing,* 2020/06/16 2020, doi: 10.1007/s11760-020-01719-y.

[19] T. E. Gilmore, F. Birgand, and K. W. Chapman, "Source and magnitude of error in an inexpensive image-based water level measurement system," (in English), *Journal of Hydrology,* vol. 496, pp. 178-186, Jul 24 2013, doi: Doi 10.1016/J.Jhydrol.2013.05.011.

[20] O. Russakovsky *et al.*, "ImageNet Large Scale Visual Recognition Challenge," *International Journal of Computer Vision,* vol. 115, no. 3, pp. 211-252, 2015/12/01 2015, doi: 10.1007/s11263-015-0816-y.

[21] H. Pham, Z. Dai, Q. Xie, M.-T. Luong, and Q. V. Le, "Meta pseudo labels," *arXiv preprint arXiv:2003.10580,* 2020.

[22] Z. Zheng, Y. Zhong, A. Ma, and L. Zhang, "FPGA: Fast Patch-Free Global Learning Framework for Fully End-to-End Hyperspectral Image Classification," *IEEE Transactions on Geoscience and Remote Sensing,* vol. 58, no. 8, pp. 5612-5626, 2020.

[23] S. W. Lo, J. H. Wu, L. C. Chen, C. H. Tseng, F. P. Lin, and C. H. Hsu, "Uncertainty Comparison of Visual Sensing in Adverse Weather Conditions," *Sensors,* vol. 16, no. 7, p. 1125, Jul 20 2016, doi: 10.3390/s16071125.

[24] J. Jiang, J. Liu, C.-Z. Qin, and D. Wang, "Extraction of Urban Waterlogging Depth from Video Images Using Transfer Learning," *Water,* vol. 10, no. 10, p. 1485, 2018, doi: 10.3390/w10101485.

[25] P. Chaudhary, S. D'Aronco, J. P. Leitão, K. Schindler, and J. D. Wegner, "Water level prediction from social media images with a multi-task ranking approach," *ISPRS Journal of Photogrammetry and Remote Sensing,* vol. 167, pp. 252-262, 2020/09/01/ 2020, doi: 10.1016/j.isprsjprs.2020.07.003.

[26] J. Jiang, C.-Z. Qin, J. Yu, C. Cheng, J. Liu, and J. Huang, "Obtaining Urban Waterlogging Depths from Video Images Using Synthetic Image Data," *Remote Sensing,* vol. 12, no. 6, p. 1014, 2020, doi: 10.3390/rs12061014.

[27] J. Jiang, J. Liu, C. Cheng, J. Huang, and A. Xue, "Automatic Estimation of Urban Waterlogging Depths from Video Images Based on Ubiquitous Reference Objects," *Remote Sensing,* vol. 11, no. 5, p. 587, 2019, doi: 10.3390/rs11050587.

[28] P. Chaudhary, S. d Aronco, M. M. D. Vitry, J. P. Leit„o, and J. D. Wegner, "FLOOD-WATER LEVEL ESTIMATION FROM SOCIAL MEDIA IMAGES," *ISPRS Annals of the Photogrammetry, Remote Sensing and Spatial Information Sciences,* vol. 4, pp. 5-12, 2019.

[29] J. Huang, J. Kang, H. Wang, Z. Wang, and T. Qiu, "A Novel Approach to Measuring Urban Waterlogging Depth from Images Based on Mask Region-Based Convolutional Neural Network," *Sustainability,* vol. 12, no. 5, p. 2149, 2020, doi: 10.3390/su12052149.

[30] M. Moy de Vitry, S. Kramer, J. D. Wegner, and J. P. Leitão, "Scalable flood level trend monitoring with surveillance cameras using a deep convolutional neural network," *Hydrol. Earth Syst. Sci.,* vol. 23, no. 11, pp. 4621-4634, 2019, doi: 10.5194/hess-23-4621-2019.

[31] S. Sarp, M. Kuzlu, M. Cetin, C. Sazara, and O. Guler, "Detecting Floodwater on Roadways from Image Data Using Mask-R-CNN," in *2020 International Conference on INnovations in Intelligent SysTems and Applications (INISTA)*, 24-26 Aug. 2020 2020, pp. 1-6, doi: 10.1109/INISTA49547.2020.9194655.

[32] M. Geetha, M. Manoj, A. S. Sarika, M. Mohan, and S. N. Rao, "Detection and estimation of the extent of flood from crowd sourced images," in *2017 International Conference on Communication and Signal Processing (ICCSP)*, 6-8 April 2017 2017, pp. 0603-0608, doi: 10.1109/ICCSP.2017.8286429.

[33] M. A. Witherow, C. Sazara, I. M. Winter-Arboleda, M. I. Elbakary, M. Cetin, and K. M. Iftekharuddin, "Floodwater detection on roadways from crowdsourced images," *Computer Methods in Biomechanics and Biomedical Engineering: Imaging & Visualization,* vol. 7, no. 5-6, pp. 529-540, 2019/11/02 2019, doi: 10.1080/21681163.2018.1488223.

[34] Y.-T. Lin, M.-D. Yang, J.-Y. Han, Y.-F. Su, and J.-H. Jang, "Quantifying Flood Water Levels Using Image-Based Volunteered Geographic Information," *Remote Sensing,* vol. 12, no. 4, p. 706, 2020, doi: doi:10.3390/rs12040706.

[35] H. Ning, Z. Li, M. E. Hodgson, and C. Wang, "Prototyping a Social Media Flooding Photo Screening System Based on Deep Learning," *ISPRS International Journal of Geo-Information,* vol. 9, no. 2, p. 104, 2020, doi: doi:10.3390/ijgi9020104.

[36] Y. Feng, C. Brenner, and M. Sester, "Flood severity mapping from Volunteered Geographic Information by interpreting water level from images containing people: A case study of Hurricane Harvey," *ISPRS Journal of Photogrammetry and Remote Sensing,* vol. 169, pp. 301-319, 2020/11/01/ 2020, doi: https://doi.org/10.1016/j.isprsjprs.2020.09.011.

[37] J. Pereira, J. Monteiro, J. Silva, J. Estima, and B. Martins, "Assessing flood severity from crowdsourced social media photos with deep neural

[37] networks," *Multimedia Tools and Applications,* vol. 79, no. 35, pp. 26197-26223, 2020/09/01 2020, doi: 10.1007/s11042-020-09196-8.
[38] A. Filonenko, Wahyono, D. C. Hernández, D. Seo, and K. Jo, "Real-time flood detection for video surveillance," in *IECON 2015 - 41st Annual Conference of the IEEE Industrial Electronics Society*, 9-12 Nov. 2015 2015, pp. 004082-004085, doi: 10.1109/IECON.2015.7392736.
[39] K. P. Menon and L. Kala, "Video surveillance system for realtime flood detection and mobile app for flood alert," in *2017 International Conference on Computing Methodologies and Communication (ICCMC)*, 18-19 July 2017 2017, pp. 515-519, doi: 10.1109/ICCMC.2017.8282518.
[40] K.-Y. Son, M. E. Yildirim, J.-S. Park, and J.-K. Song, *Flood detection by using FCN-AlexNet* (Eleventh International Conference on Machine Vision (ICMV 2018)). SPIE, 2019.
[41] T. Oga, R. Harakawa, S. Minewaki, Y. Umeki, Y. Matsuda, and M. Iwahashi, "River state classification combining patch-based processing and CNN," *PLOS ONE,* vol. 15, no. 12, p. e0243073, 2020, doi: 10.1371/journal.pone.0243073.
[42] N. H. Jafari, X. Li, Q. Chen, C.-Y. Le, L. P. Betzer, and Y. Liang, "Real-time water level monitoring using live cameras and computer vision techniques," *Computers & Geosciences,* vol. 147, p. 104642, 2021/02/01/ 2021, doi: https://doi.org/10.1016/j.cageo.2020.104642.
[43] O. Ronneberger, P. Fischer, and T. Brox, "U-Net: Convolutional Networks for Biomedical Image Segmentation," Cham, 2015: Springer International Publishing, in Medical Image Computing and Computer-Assisted Intervention – MICCAI 2015, pp. 234-241.
[44] M. Raghu, C. Zhang, J. Kleinberg, and S. Bengio, "Transfusion: Understanding transfer learning for medical imaging," in *Advances in neural information processing systems*, 2019, pp. 3347-3357.
[45] R. Jain, P. Nagrath, G. Kataria, V. S. Kaushik, and D. J. Hemanth, "Pneumonia detection in chest X-ray images using convolutional neural networks and transfer learning," (in English), *Measurement,* Article vol. 165, p. 10, Dec 2020, Art no. 108046, doi: 10.1016/j.measurement.2020.108046.
[46] B. Neyshabur, H. Sedghi, and C. Zhang, "What is being transferred in transfer learning?," *Advances in Neural Information Processing Systems,* vol. 33, 2020.
[47] A. G. Howard *et al.*, "Mobilenets: Efficient convolutional neural networks for mobile vision applications," *arXiv preprint arXiv:1704.04861,* 2017.
[48] S. Bianco, R. Cadene, L. Celona, and P. Napoletano, "Benchmark Analysis of Representative Deep Neural Network Architectures," *IEEE Access,* vol. 6, pp. 64270-64277, 2018, doi: 10.1109/ACCESS.2018.2877890.
[49] D. P. Kingma and J. Ba, "Adam: A method for stochastic optimization," *arXiv preprint arXiv:1412.6980,* 2014.
[50] R. R. Selvaraju, M. Cogswell, A. Das, R. Vedantam, D. Parikh, and D. Batra, "Grad-cam: Visual explanations from deep networks via gradient-based localization," in *Proceedings of the IEEE international conference on computer vision*, 2017, pp. 618-626.
[51] B. Zhou, A. Khosla, A. Lapedriza, A. Oliva, and A. Torralba, "Learning deep features for discriminative localization," in *Proceedings of the IEEE conference on computer vision and pattern recognition*, 2016, pp. 2921-2929.
[52] J. Redmon and A. Farhadi, "Yolov3: An incremental improvement," *Computer Vision and Patern Recognition (CVPR),* pp. 126-134, 2018.
[53] K. He, G. Gkioxari, P. Dollár, and R. Girshick, "Mask r-cnn," in *Proceedings of the IEEE international conference on computer vision*, 2017, pp. 2961-2969.
[54] "Line (software)." Wikipedia contributors. https://en.wikipedia.org/w/index.php?title=Line_(software)&oldid=983530597 (accessed 6 November 2020, 2020).
[55] "Digital Report 2020 Taiwan." Taiwan Network Information Center. https://datareportal.com/reports/digital-2020-taiwan (accessed 6 November 2020, 2020).